\documentclass[letterpaper, 10 pt, conference]{ieeeconf}  

\IEEEoverridecommandlockouts                              

\usepackage{graphicx}
\graphicspath{{./figs/}}
\usepackage{amsmath}
\usepackage{mathtools}

\usepackage{etex} 

\usepackage[utf8]{inputenc}
\usepackage[T1]{fontenc}
\usepackage{fixltx2e}
\usepackage{longtable}
\usepackage{float}
\usepackage{wrapfig}
\usepackage{soul}
\usepackage{textcomp}
\usepackage{marvosym}
\usepackage{wasysym}
\usepackage{latexsym}
\usepackage[table]{xcolor}
\usepackage{dcolumn}
\usepackage{booktabs}
\usepackage{subcaption} 
\usepackage[labelformat=simple]{subcaption}

\usepackage{enumerate} 

\usepackage{graphicx}
\usepackage{tikz}
\usetikzlibrary{decorations.fractals}

\usepackage{empheq}
\usepackage{bm} 
\usepackage{movie15}

\usepackage{amsfonts}
\usepackage{amscd}
\usepackage{amssymb}

\usepackage[export]{adjustbox}

\usepackage[d]{esvect}
\usepackage{manfnt} 

\usepackage{xcolor}

\newcommand{\Vect}[1]{%
  \mathbf{#1}
}

\newcommand{\norm}[1]{\left\lVert#1\right\rVert}

\usepackage{amsmath,mleftright}
\usepackage{xparse}
\NewDocumentCommand{\evalat}{sO{\big}mm}{%
  \IfBooleanTF{#1}
   {\mleft. #3 \mright|_{#4}}
   {#3#2|_{#4}}%
}

\makeatletter
\let\NAT@parse\undefined
\makeatother
\usepackage[hidelinks]{hyperref}

\title{\LARGE \bf
Trajectory optimization for a class of robots belonging to Constrained Collaborative Mobile Agents (CCMA) family
}

\author{Nitish Kumar$^{*}$, Stelian Coros
\thanks{Nitish Kumar and Stelian Coros are with the Computational Robotics Lab in the Institute for Intelligent Interactive Systems (IIIS), ETH Zurich Switzerland. $\{$nitish.kumar@inf.ethz.ch, stelian.coros@inf.ethz.ch$\}$}
\thanks{* corresponding author}
}

\begin{document}
\bstctlcite{MyBSTcontrol}

\maketitle
\thispagestyle{empty}
\pagestyle{empty}

\begin{abstract}
We present a novel class of robots belonging to Constrained Collaborative Mobile Agents (CCMA) family which consists of ground mobile bases with non-holonomic constraints. Moreover, these mobile robots are constrained by closed-loop kinematic chains consisting of revolute joints which can be either passive or actuated. We also describe a novel trajectory optimization method which is general with respect to number of mobile robots, topology of the closed-loop kinematic chains and placement of the actuators at the revolute joints. We also extend the standalone trajectory optimization method to optimize concurrently the design parameters and the control policy. We describe various CCMA system examples, in simulation, differing in design, topology, number of mobile robots and actuation space. The simulation results for standalone trajectory optimization with fixed design parameters is presented for CCMA system examples. We also show how this method can be used for tasks other than end-effector positioning such as internal collision avoidance and external obstacle avoidance. The concurrent design and control policy optimization is demonstrated, in simulations, to increase the CCMA system workspace and manipulation capabilities.  Finally, the trajectory optimization method is validated in experiments through two $4$-DOF prototypes consisting of $3$ tracked mobile bases.
\end{abstract}


\section{Introduction}

\subsection{Motivation}
Mechanisms are fundamental to address problems of robotic manipulation involving motion generation and force transmission. Examples include serial mechanisms such as industrial manipulators and closed-loop kinematic mechanisms such as pick and place delta robots or cable driven robots. Different types of mobile platforms such as aerial quadcopters, ground mobile robots (omni-directional, tracked, wheeled) and legged robots provide solutions to the problem of robotic mobility. Robotic mobility and robotic manipulation basically seek motion generation and force output capability at global and local level, respectively. Several important applications such as field robotics, construction robotics~\cite{loveridge_robots_2017, hack_digital_2017, kumar_design_2017, noauthor_digital_2018, hack_mesh_2017}, service robotics require robots to be mobile as wells as able to do manipulation tasks. Interfacing a single mobile base with a serial manipulator~\cite{giftthaler_mobile_2017,keating_toward_2017} or multiple mobile bases with closed-loop kinematic chains~\cite{kumar_optimization_2019_iros} is a potential solution to solving both the challenges of global mobility and local manipulation. Instead of idealizing such systems as monolithic robots designed once in its lifetime for a particular task, our long term goal is to investigate them within a much broader abstraction of heterogeneous, modular, mobile, multi-robot and reconfigurable systems. Our vision is a ubiquitous robot ecosystem which is highly mobile, modular,  customizable with heterogeneous components and
readily operational for varied tasks.

\begin{figure}[!t] 
  \centering
  \vspace{7pt}
    \includegraphics[width=0.95\columnwidth]{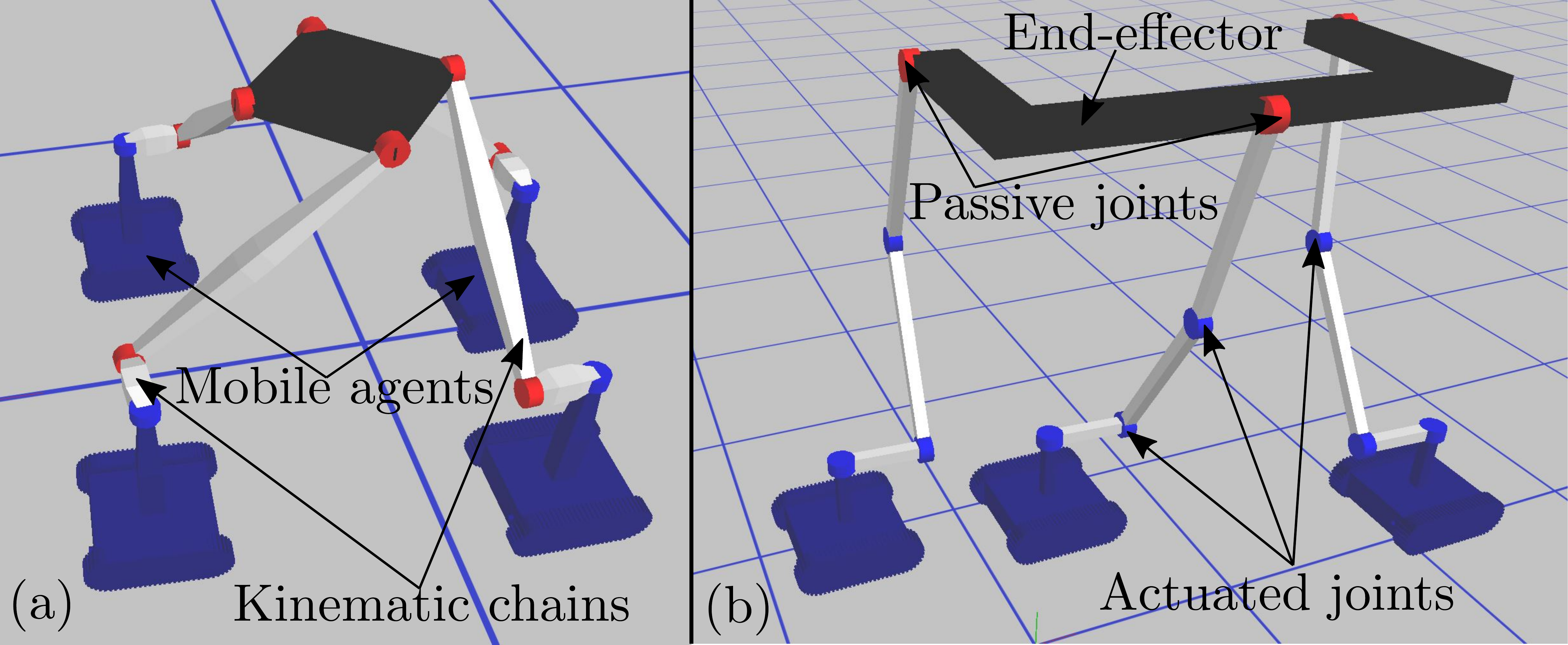}
    \caption{Constrained collaborative mobile agents concept: a number of mobile bases $0,1\cdots n_m-1$ constrained with active-passive closed-loop kinematic chains (gray bars connected with red passive joints and blue actuated joints) manipulating an end-effector (black polygon).}
    \label{fig:concepCCMA}
\end{figure}

The idea of interfacing multiple mobile bases with closed-loop kinematic chains presents a huge potential to exploit the large design space for task-based customization and reconfiguration of modular mobile multi-robot systems. 
Moreover, if a subset of joints in the closed-loop kinematic chains are actuated, a large control space can be generated for active-passive closed-loop kinematic chains. This could especially be useful in optimizing the stiffness or payload capacity of the system. In our previous work~\cite{kumar_optimization_2019_iros}, we introduced the concept of Constrained Collaborative Mobile Agents (CCMA). A class of robots consisting of omni-directional mobile bases and fully passive closed-loop kinematic chains was described. In this paper, we are continuing this work and introduce another novel class of robots belonging to CCMA family (Fig.~\ref{fig:concepCCMA}), which has non-holonomic mobile bases and active-passive closed-loop kinematic chains with revolute joints.

The motion generation and force transmission are two fundamental capabilities which we would like to investigate with the CCMA concept, in the long term. The focus of this paper is, however, on the motion generation capabilities of the CCMA systems. In this respect, one of the most important tasks of a mobile robotic system is the gross manipulation or gross positioning of a more precise robotic end-effector. For mobile bases with non-holonomic constraints, this task is non-trivial and requires trajectory optimization techniques. Even for mobile bases with holonomic constraints, certain tasks such as external obstacle avoidance require trajectory optimization techniques. Moreover, the closed-loop kinematic chains along with the actuation of a subset of joints impose even more kinematic constraints on the motion of the mobile agents. Therefore, we develop and present a novel framework for kinematic motion planning and trajectory optimization for the class of the robots presented in this paper. We also extend this trajectory optimization method to allow design optimization concurrently. This showcases the ability of the framework to adapt designs of the robots for different motion generation requirements for different tasks. 

\subsection{Related work}
There are several specific instances of multiple mobile bases being used for performing collaborative tasks, in the literature. An architectural scale installation was built through collaboration between multiple aerial quadcopters in the work~\cite{augugliaro_flight_2014}. A mobile cable driven system using  multiple mobile bases was demonstrated in the work ~\cite{rasheed_kinematic_2018,zi_localization_2015} for logistics applications. Mobile parallel robots with multiple omni-directional mobile bases was presented in the work~\cite{wan_design_2010, hu_singularity_2009, hu_type_2009}. In this paper, we present a novel class of multi-robot collaborative systems, which has multiple non-holonomic wheeled mobile bases constrained with active-passive closed-loop kinematics chains. 

Interesting examples of mobile robotic systems with fully actuated legs, and wheels with non-holonomic constraints have been proposed in the work~\cite{giftthaler_efficient_2017, giordano_kinematic_2009, geilinger_skaterbots:_2018}. In the current work, the closed-loop kinematic chains in the CCMA system examples need not be fully actuated. Therefore, the class of robots introduced in the paper exhibit a vast design and actuation space with fully passive to fully actuated revolute joints.

An overview of planning methods for robotic systems with non-holonomic constraints can be found in the work~\cite{lavalle_planning_2006,laumond_guidelines_1998,jean_control_2014}.
The trajectory optimization and kinematic motion planning method presented in the current paper complements other optimal control approaches~\cite{giftthaler_efficient_2017, giordano_kinematic_2009, dietrich_singularity_2011, geilinger_skaterbots:_2018} and sampling based approaches~\cite{yakey_randomized_2001, mcmahon_sampling-based_2018,
oriolo_motion_2005}. However none of these optimal control or sampling based approaches for trajectory optimization or motion planning discuss the possibility to do concurrent design and control policy optimization over a trajectory. In our approach, the design parameters of the mobile multi-robot system can be optimized and updated along with control policy during the trajectory optimization step. This could specially be powerful when the process or task itself is evolving during initial conception stages and could require several design iterations of a modular, mobile, multi-robot system. 
In our method, rigid body kinematics is modeled on a constraint based formulation presented in the paper~\cite{coros_computational_2013,thomaszewski_computational_2014}, which abstracts a rigid body robotic system to a collection of rigid bodies connected with kinematic constraints imposed by the joints. For our trajectory optimization technique, we calculate the derivatives of the kinematic constraints analytically, required for the gradient-based methods (e.g. L-BFGS, Gauss-Newton). We utilize the sensitivity analysis techniques~\cite{mcnamara_fluid_2004, auzinger_computational_2018,cao_adjoint_2002, jackson_second-order_1988, zimmermann_puppetmaster:_2019, Coros-RSS-19} to calculate the derivatives of the tasks formulated as objective functions with respect to control and design parameters of the CCMA system. 

\subsection{Contributions}
In this paper, contributions are two-fold. First contribution is the development of a novel hardware architecture consisting of multiple mobile agents, with non-holonomic constraints, further constrained by active-passive closed-loop kinematic chains manipulating an end-effector. We present, in simulation, several designs ranging in design parameters, degrees of freedom (DOF), morphology, number of mobile agents and number of actuated joints showing the potential for scalability, task-adaptability and reconfigurability.

To exploit the aspects of modularity, design and control flexibility, and collaborative manipulation, an optimization framework for CCMA system is essential for simulation and trajectory optimization. 
The second contribution is the development of this optimization framework for concurrent design and trajectory optimization, and kinematic motion planning of the CCMA system. Our optimization framework is independent of design parameters, DOF, morphology, number of mobile agents and number of actuated joints in a CCMA system. We further evaluate this optimization framework on several CCMA system examples and demonstrate several tasks ranging from gross end-effector positioning, internal mobile agents collision avoidance to external obstacle collision avoidance. 
\subsection{Paper organization}
We present the modeling of the specific instantation of the CCMA concept with non-holonomic constraints in Sec.~\ref{sec:CCMASystemModeling}. The optimization framework for the concurrent design and trajectory optimization, and kinematic motion planning is presented in Sec.~\ref{sec:OptFramework}. In Sec.~\ref{sec:simResults}, we present and discuss the simulation results which evaluate the optimization framework on different CCMA system examples for different objectives. The results of trajectory optimization method on two physical prototypes is discussed in Sec.~\ref{sec:expResults}.
In Sec.~\ref{sec:final-sec}, we present the conclusions and directions for the future work. 

\section{Constrained collaborative mobile agents system - description and modeling}
\begin{figure}[!t] 
  \centering
  \vspace{7pt}
    \includegraphics[width=0.9\columnwidth]{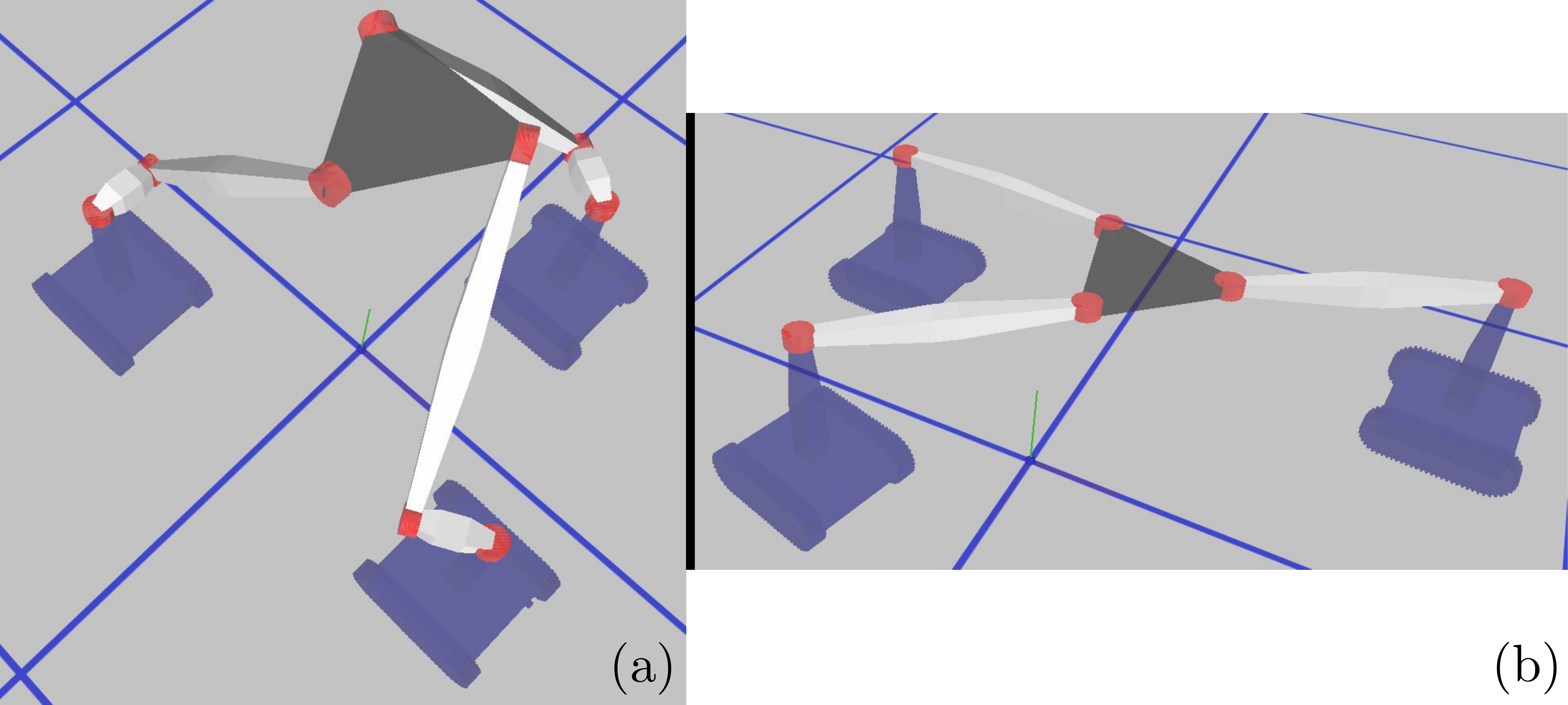}
    \caption{An example (a) 4-DOF and (b) 3-DOF CCMA system consisting of tracked mobile robots with non-holonomic constraints in simulation. All joints are passive in this example.}
    \label{fig:CCMAexamples}
\end{figure}

The Fig.~\ref{fig:concepCCMA} illustrates an instantiation of the CCMA concept presented in this paper. It consists of active-passive closed-loop kinematic chains connecting the end-effector  to a number of mobile bases with non-holonomic constraints. The mobile agents along with a set of actuated joints in the closed-loop kinematic chains, form the actuators in the system to control the end-effector. The mobile agents (wheeled or tracked systems with non-holonomic constraints) have $2$-DOF each. Fig.~\ref{fig:concepCCMA}(a) shows a system example with $4$ mobile bases and active kinematic chains consisting of one actuated joint each.  Fig.~\ref{fig:concepCCMA}(b) shows a system example with $3$ mobile bases and active kinematic chains consisting of three actuated joints each. Two other examples of this hardware architecture are shown in Fig.~\ref{fig:CCMAexamples}, differing in design and morphology. However, in these two examples, the kinematic chains are totally passive, as compared to  the system examples in Fig.~\ref{fig:concepCCMA} where a subset of revolute joints are actuated.

\label{sec:CCMASystemModeling}


\subsection{Notations and preliminaries}
\begin{description}
\item[$n_b$] total number of rigid bodies in CCMA
system.
\item[$n_m$] number of mobile robots.
\item[$n_{a}$] number of actuated joints.
\item[$n_{t}$] number of states in the trajectory.
\item[$\Vect{st_j}$] intermediate state vector of the system at time $t_j$ which has a size of $6 \cdot n_b$, where $j=0, 1, 2, \cdots n_t$.
\item[$\Vect{u_j}$]  intermediate control vector which takes the state $\Vect{s_j}$ to $\Vect{s_{j+1}}$ in time step $\delta t$. The size of $\Vect{u_j}$ is $2 \cdot n_m + n_{a}$. It is formed by stacking the linear and angular speed of all mobile bases followed by angular speed of all actuated rotary joints.
\item[$\Vect{a_j}$]  it is formed by stacking the x, y components of the linear velocity and, angular speed of the mobile bases followed by angular speeds of the actuated rotary joints. The size of $\Vect{a_j}$ is $3 \cdot n_m + n_{a}$. 
\item[$\Vect{m_j}$] it contains the absolute values of the mobile base poses and rotary actuator angles. Specifying this vector also specifies the state $\Vect{s_j}$ of CCMA system. The size of $\Vect{m_j}$ is $3 \cdot n_m + n_{a}$.
\item[$\Vect{dp}$] vector of design parameters. In this paper, we consider the kinematic parameters, affecting geometry of the rigid bodies, which have impact on the motion generation capability and workspace.  
\item[$\Vect{C_j}$] a vector of constraints, corresponding to state $\Vect{st_j}$, which include the constraints output by (a) each passive joint (b) kinematic constraints which correspond to the planar constraint on the mobile robot (c) mobile robot actuator constraints assuming two motorized prismatic actuators along $\mathbf{x_g}$, $\mathbf{y_g}$ and a rotary actuator along $\mathbf{z_g}$. ($\Vect{O_g}$, $\mathbf{x_g}$, $\mathbf{y_g}$, $\mathbf{z_g}$) represent world reference frame with origin at $\Vect{O_g}$ and three unit vectors along $\mathbf{x_g}$, $\mathbf{y_g}$, $\mathbf{z_g}$ d) each rotary actuator which are obtained by fixing the value of angle in the motorized joint. The mathematical modeling of the constraints for (a), (b) and (c) is described in our previous paper~\cite{kumar_optimization_2019_iros}.
\end{description}
$\Vect{st}$, $\Vect{u}$, $\Vect{a}$ and $\Vect{m}$ are trajectory vectors obtained after stacking the vectors $\Vect{st_j}$, $\Vect{u_j}$, $\Vect{a_j}$ and $\Vect{m_j}$, respectively.

\begin{figure}[!t] 
  \centering
  \vspace{7pt}
    \includegraphics[width=0.95\columnwidth]{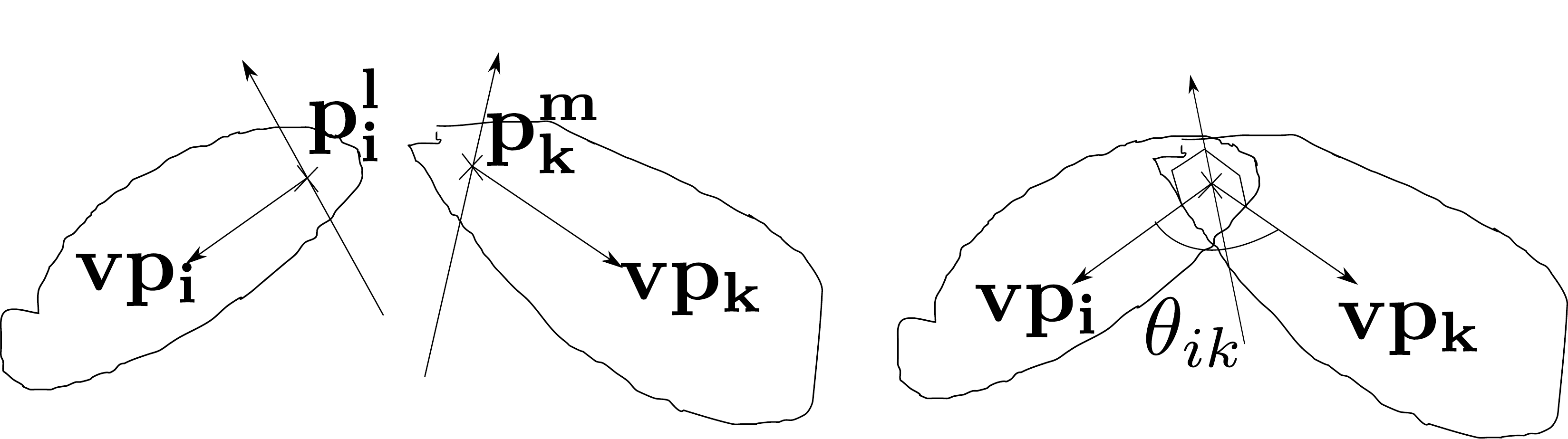}
    \caption{Two rigid bodies $i$ and $k$ connected by a revolute joint connection and a rotary actuator fixing the value of $\theta_{ik}$.}
    \label{fig:rvCon}
\end{figure}

\subsubsection{Rotary actuator constraints between two rigid bodies}
A rigid body $i$ has $6$-DOF which is described by its state $\Vect{s_i} = [\gamma_i, \beta_i, \alpha_i, \Vect{T_i}]$. $\Vect{s_i}$ consists of three Euler angles and translation vector $\Vect{T_i} = [x_i, y_i, z_i]$ from global reference frame to rigid body local co-ordinate system. Thus any point $\Vect{\bar{p}_i}$ or free vector $\Vect{\bar{v}_i}$ expressed in local co-ordinate system of a rigid body $i$ can be converted to global world co-ordinates as $\Vect{p_i} = \Vect{R_{\gamma_i}} \cdot \Vect{R_{\beta_i}} \cdot \Vect{R_{\alpha_i}} \cdot \Vect{\bar{p}_i} +  \Vect{T_i}$ or $\Vect{v_i} = \Vect{R_{\gamma_i}} \cdot \Vect{R_{\beta_i}} \cdot \Vect{R_{\alpha_i}} \cdot \Vect{\bar{v}_i} $. $\Vect{R}$ is an elementary $3 \times 3$ rotation matrix.

Let $\Vect{\bar{vp}_i}$ and $\Vect{\bar{vp}_k}$ be two unit vectors in the plane perpendicular to the rotation axis and attached to rigid body $i$ and $k$ at points $\Vect{\bar{p}^{l}_i}$ and $\Vect{\bar{p}^{m}_k}$ (Fig.~\ref{fig:rvCon}), respectively.  Let $\theta_{ik}$ be angle between $\Vect{vp_i(\bar{vp}_i)}$ and $\Vect{vp_k(\bar{vp}_k)}$ about the common rotation axis. Fixing the angle of rotary actuator by $\theta_{ik}$ imposes the constraint $\Vect{C^{a}_{ik}}=[\Vect{R_{\gamma^m_i}} \cdot \Vect{R_{\beta^m_i}} \cdot \Vect{R_{\alpha^m_i}}\cdot (\Vect{R_{\theta_{ik}}}\cdot \Vect{\bar{vp}_i}) - \Vect{R_{\gamma^m_k}} \cdot \Vect{R_{\beta^m_k}} \cdot \Vect{R_{\alpha^m_k}}\cdot \Vect{\bar{vp}_k}]$. 
\subsubsection{Design parameters $\Vect{dp}$}
Each rigid body $i$ geometry is defined by a number of points, such as the $l^{th}$ point $\Vect{\bar{p}^{l}_i}$ on it (Fig.~\ref{fig:rvCon}). Let $\Vect{\bar{po}^{l}_i}  = \Vect{\bar{p}^{l}_i}$ before the start of the design optimization.
A design parameter $dp^l_i$ is defined to modify every such point on the rigid body as follows: $\Vect{\bar{p}^{l}_i} = \Vect{\bar{po}^{l}_i} + dp^l_i \cdot \frac{\Vect{\bar{po}^{l}_i}}{\norm{\Vect{\bar{po}^{l}_i}}}$. The continuous scalar variable $dp^l_i$ can take both positive and negative values. Therefore, $\Vect{dp}$ accumulates these design parameters for every defined point on each rigid body in the CCMA system. These parameters can implicitly define the length of a link or geometry of a polygonal end-effector or support polygon of a mobile base, in a CCMA system. 

\section{Optimization framework}
\label{sec:OptFramework}
The optimization framework, and the trajectory optimization algorithm  consists of two steps.
\subsection{Forward simulation of the system state trajectory $\Vect{st}$ for given control input $\Vect{u}$ and design parameters $\Vect{dp}$}
Following schematic shows the system state trajectory evolution, when control inputs $\Vect{u_{j}}, j=0, 1, 2, 3 \cdots n_t-2$ is successively applied starting from state $\Vect{st_0}$.
\begin{equation*}
\underset{\underset{\Vect{st_0}}{\Big\downarrow}}{\Vect{m_0}} 
\xrightarrow[]{\Vect{u_0}}
\underset{\underset{\Vect{st_1}}{\Big\downarrow}}{\Vect{m_1}} 
\xrightarrow[]{\Vect{u_1}}
\underset{\underset{\Vect{st_2}}{\Big\downarrow}}{\Vect{m_2}} 
\cdots
\underset{\underset{\Vect{st_j}}{\Big\downarrow}}{\Vect{m_j}} 
\xrightarrow[]{\Vect{u_j}}
\cdots
\underset{\underset{\Vect{st_{n_t-2}}}{\Big\downarrow}}{\Vect{m_{n_t-2}}} 
\xrightarrow[]{\Vect{u_{n_t-2}}}
\underset{\underset{\Vect{st_{n_t-1}}}{\Big\downarrow}}{\Vect{m_{n_t-1}}} 
\end{equation*}

In order to calculate the next state $\Vect{st_{j+1}}$ from a starting state $\Vect{st_{j}}$ and control input $\Vect{u_{j}}$, we solve an optimization problem and minimize an energy $E_j$ which is sum of the mechanism constraint violations as below:
\begin{equation}
\begin{aligned}
& \Vect{st_{j+1}} = \underset{\Vect{st}}{\text{arg min}}
& & [{E_j}(\Vect{m_j}, \Vect{st}, \Vect{u_j}, \Vect{dp}) = \frac{1}{2}\Vect{C_j}^{T}\Vect{C_j}]\\
\end{aligned}
\label{eqn:nextState}
\end{equation}

For mobile base $k=0,1,2\cdots (n_m-1)$ with non-holonomic constraints:
\begin{align}
\begin{split}
&\Vect{u_j}[2k] = \text{linear velocity of the mobile robot $k$ in state $\Vect{st_j}$}\\
&\Vect{u_j}[2k+1] = \text{angular velocity of the mobile robot $k$ in $\Vect{st_j}$}\\
&\Vect{a_j}[3k+1] = \text{component of the linear velocity along $\Vect{x_g}$}\\
&\Vect{a_j}[3k+2] = \text{component of the linear velocity along $\Vect{y_g}$}\\
&\theta^m_{j,k} = \Vect{m_j}[3k] =  \text{global orientation of the mobile robot}\\
&\Vect{a_j}[3k]=\Vect{u_j}[2k]; \qquad \Vect{a_j}[3k+1]=\Vect{u_j}[2k+1]\cdot \text{cos}(\theta^m_{j,k})\\
&\Vect{a_j}[3k+2]=\Vect{u_j}[2k+1]\cdot \text{sin}(\theta^m_{j,k})\\
& \Vect{a_j} = f_j(\Vect{u_j}, \Vect{m_j})
\end{split}
\label{eqn:implicitNH}
\end{align}
Eqn.~\ref{eqn:implicitNH} implicitly takes into account the non-holonomic constraints while updating $\Vect{a_j}$ from the $\Vect{u_j}$ and $\Vect{m_j}$. Therefore, the non-holonomic constraints are always satisfied within the forward simulation step itself. They are never optimized as part of the trajectory optimization step which is described in the next subsection.

The state transition Eqn.~\ref{eqn:nextState} can also be written as a function of $\Vect{m_{j+1}}$ as below:
\begin{equation}
\begin{aligned}
&\Vect{m_{j+1}}=\Vect{m_{j}}+f_j(\Vect{u_j}, \Vect{m_j})\cdot \delta t\\
& \Vect{st_{j+1}} = \underset{\Vect{st}}{\text{arg min}} \quad
 {E_j}(\Vect{m_{j+1}}, \Vect{st}, \Vect{dp}) \\
\end{aligned}
\label{eqn:nextState2}
\end{equation}

\subsection{Trajectory optimization leading to calculation of the control input evolution $\Vect{u}$ and design parameters $\Vect{dp}$}
We solve an optimization problem where an objective function $\mathcal{O}$ is evaluated over the entire trajectory, and minimized to calculate the control inputs $ \Vect{u_0},\Vect{u_1} \cdots \Vect{u_j} \cdots \Vect{u_{n_t-2}}$ and design parameters $\Vect{dp}$. The design parameters $\Vect{dp}$ remain the same when successive control inputs $ \Vect{u_0},\Vect{u_1} \cdots \Vect{u_j} \cdots \Vect{u_{n_t-2}}$ are applied and they are updated along with $\Vect{u}$ at start of each optimization step.

The assumption of the sensitivity analysis (Sec.~\ref{sec:sensitivityAnalysis}) states that as result of the optimization upon convergence, in Eqn.~\ref{eqn:nextState2}, we get gradient of energy function $E_j$, $\Vect{g_j}(\Vect{m_j}, \Vect{st_{j+1}}, \Vect{u_j}, \Vect{dp})=0$ for all $\Vect{u_j}$ and $\Vect{dp}$. This does not ensure, however, that the residual constraint energy $E_{rj}=E_j(\Vect{m_{j+1}}, \Vect{st_{j+1}}, \Vect{dp})$ after optimization, from the forward simulation step, will be zero. Therefore, we add the residual constraint energies to the objective function $\mathcal{O}$ to make sure that the remaining residual constraint energies also go to zero during the trajectory optimization.
\begin{multline}
\underset{\Vect{u}, \Vect{dp}}
{\text{arg min}} \quad
 \mathcal{O}(\Vect{st},\Vect{u}, \Vect{m}, \Vect{dp})
= \\
\sum_{i=0}^{n_{task}-1}w_i\cdot\mathcal{O}_{task_i} (\Vect{st},\Vect{u}, \Vect{m}, \Vect{dp}) \quad + \\
\lambda_E\cdot\sum_{j=0}^{n_t-1} E_j(\Vect{m_{j+1}}, \Vect{st_{j+1}}, \Vect{dp})
\label{eqn:objectiveFunction2}
\end{multline}
This objective function $\mathcal{O}$, in addition to the task of reaching a specified end state can consist of other tasks such as external obstacle avoidance, internal link collision avoidance and internal mobile  agent collision avoidance.
In order to solve this optimization problem through a gradient based minimization approach, we need calculation of the gradients of the objection function $\frac{d\mathcal{O}}{d\Vect{u_i}}, i=0,1,2,3 \cdots n_t-2$ and $\frac{d\mathcal{O}}{d\Vect{dp}}$, as described in Eqn~\ref{eqn:finalGradient}. The parameters $\Vect{m}$ and $\Vect{dp}$ are independent of each other. 
\begin{equation}
\begin{aligned}
&\frac{d\mathcal{O}}{d\Vect{u_i}}=\frac{\partial\mathcal{O}}{\partial\Vect{u_i}} + \sum_{j=0}^{n_t-1} \frac{\partial\mathcal{O}}{\partial\Vect{st_j}}\cdot\frac{d\Vect{st_j}}{d\Vect{u_i}} + 
\sum_{j=0}^{n_t-1} \frac{\partial\mathcal{O}}{\partial\Vect{m_j}}\cdot\frac{d\Vect{m_j}}{d\Vect{u_i}}\\
&\frac{d\mathcal{O}}{d\Vect{dp}}=\frac{\partial\mathcal{O}}{\partial\Vect{dp}} + \sum_{j=0}^{n_t-1} \frac{\partial\mathcal{O}}{\partial\Vect{st_j}}\cdot\frac{d\Vect{st_j}}{d\Vect{dp}}
\end{aligned}
\label{eqn:finalGradient}
\end{equation}
However, calculation of the sensitivities $\frac{d\Vect{st_j}}{d\Vect{u_i}}$, $\frac{d\Vect{m_j}}{d\Vect{u_i}}$ and $\frac{d\Vect{st_j}}{d\Vect{dp}}$  is not possible directly in closed form and is computationally prohibitive using finite differences. 
Therefore, we make use of the sensitivity analysis to indirectly calculate them analytically.
\subsubsection{Sensitivity analysis}
\label{sec:sensitivityAnalysis}
Since control vector $\Vect{u_k},\quad k=0,1,2 \cdots n_t-2$ only affects the future states from $\Vect{st_{k+1}}$ onwards, we can state that $\frac{d\Vect{st_j}}{d\Vect{u_k}} = 0; \frac{d\Vect{m_j}}{d\Vect{u_k}} = 0 \quad \forall j \leq k$.
As result of the optimization upon convergence, in Eqn.~\ref{eqn:nextState}, we get gradient of energy function $E_j$, $\Vect{g_j}(\Vect{m_j}, \Vect{st_{j+1}}, \Vect{u_j}, \Vect{dp})=0$ for all $\Vect{u_j}$. The parameters $\Vect{u}$ and $\Vect{dp}$ are independent of each other. 
\begin{align*}
& \frac{\partial\Vect{g_j}}{\partial\Vect{u_j}} + \frac{\partial\Vect{g_j}}{\partial\Vect{m_j}}\cdot\frac{d\Vect{m_j}}{d\Vect{u_j}}
+ \frac{\partial\Vect{g_j}}{\partial\Vect{st_{j+1}}}\cdot\frac{d\Vect{st_{j+1}}}{d\Vect{u_j}} = 0 \implies \\
& \frac{\partial\Vect{g_j}}{\partial\Vect{u_j}} + \frac{\partial\Vect{g_j}}{\partial\Vect{st_{j+1}}}\cdot\frac{d\Vect{st_{j+1}}}{d\Vect{u_j}} = 0  \implies\\
& \frac{d\Vect{st_{j+1}}}{d\Vect{u_j}} = -\Bigg(\frac{\partial\Vect{g_j}}{\partial\Vect{st_{j+1}}}\Bigg)^{-1}\cdot\frac{\partial\Vect{g_j}}{\partial\Vect{u_j}}\\
& \frac{d\Vect{st_{j+1}}}{d\Vect{u_j}}=
\frac{\partial\Vect{st_{j+1}}}{\partial\Vect{u_j}} + \frac{\partial\Vect{st_{j+1}}}{\partial\Vect{m_j}}\cdot\frac{d\Vect{m_j}}{d\Vect{u_j}} = \frac{\partial\Vect{st_{j+1}}}{\partial\Vect{u_j}}\\
& \frac{d\Vect{st_{j+1}}}{d\Vect{u_j}} = \frac{\partial\Vect{st_{j+1}}}{\partial\Vect{a_j}}\cdot \frac{\partial\Vect{a_{j}}}{\partial\Vect{u_j}}  \\ 
& \text{it can be shown, }
\frac{\partial\Vect{st_{j+1}}}{\partial\Vect{a_j}} = -\Bigg(\frac{\partial\Vect{g_j}}{\partial\Vect{st_{j+1}}}\Bigg)^{-1}\cdot\frac{\partial\Vect{g_j}}{\partial\Vect{a_j}}
\end{align*}

Similarly, gradient of energy function $E_j$, $\Vect{g_j}(\Vect{m_j}, \Vect{st_{j+1}}, \Vect{u_j}, \Vect{dp})=0$ for all $\Vect{dp}$.
\begin{align*}
& \frac{\partial\Vect{g_j}}{\partial\Vect{dp}} + \frac{\partial\Vect{g_j}}{\partial\Vect{m_j}}\cdot\frac{d\Vect{m_j}}{d\Vect{dp}}
+ \frac{\partial\Vect{g_j}}{\partial\Vect{st_{j+1}}}\cdot\frac{d\Vect{st_{j+1}}}{d\Vect{dp}} = 0 \implies \\
& \frac{\partial\Vect{g_j}}{\partial\Vect{dp}} + \frac{\partial\Vect{g_j}}{\partial\Vect{st_{j+1}}}\cdot\frac{d\Vect{st_{j+1}}}{d\Vect{dp}} = 0  \implies\\
& \frac{d\Vect{st_{j+1}}}{d\Vect{dp}} = -\Bigg(\frac{\partial\Vect{g_j}}{\partial\Vect{st_{j+1}}}\Bigg)^{-1}\cdot\frac{\partial\Vect{g_j}}{\partial\Vect{dp}}
\end{align*}

$\frac{\partial\Vect{g_j}}{\partial\Vect{st_{j+1}}}$ is the hessian of the energy $E_j$ evaluated at $st_{j+1}$. $\frac{\partial\Vect{g_j}}{\partial\Vect{a_j}}$, $\frac{\partial\Vect{g_j}}{\partial\Vect{u_j}}$ and $\frac{\partial\Vect{g_j}}{\partial\Vect{dp}}$ is how the gradient of the energy $E_j$ changes with change in the input $\Vect{a_j}$,
$\Vect{u_j}$, $\Vect{dp}$ only, respectively. These terms are calculated analytically. First all sensitivities $\frac{d\Vect{st_{j+1}}}{d\Vect{u_j}}, j=0, 1,2,3 \cdots n_t-2$ are calculated.

\subsubsection{Recurrence relation}

To calculate rest of the sensitivities $\frac{d\Vect{st_{j}}}{d\Vect{u_k}},\frac{d\Vect{m_{j}}}{d\Vect{u_k}} \quad \text{for} \quad k+1<j \leq n_t-1$; $k=0,1,2, \cdots n_t-2$, we derive and present the following recurrence relation:
\begin{align}
\begin{split}
&\frac{\partial\Vect{m_{j+1}}}{\partial\Vect{a_{j}}} = \Vect{I}\cdot \delta t \quad \text{from Eqn.~\ref{eqn:nextState2}}\\
&\frac{\partial\Vect{st_{j+1}}}{\partial\Vect{m_{j+1}}}\cdot \delta t = \frac{\partial\Vect{st_{j+1}}}{\partial\Vect{m_{j+1}}} \cdot \frac{\partial\Vect{m_{j+1}}}{\partial\Vect{a_{j}}}
= \frac{\partial\Vect{st_{j+1}}}{\partial\Vect{a_{j}}}\\
&\frac{\partial\Vect{st_{j+1}}}{\partial\Vect{u_k}} = 0 \quad \forall \quad k\neq j\\
&\frac{d\Vect{st_{j+1}}}{d\Vect{u_k}} =  \frac{\partial\Vect{st_{j+1}}}{\partial\Vect{u_{k}}}+\frac{\partial\Vect{st_{j+1}}}{\partial\Vect{m_{j+1}}}\cdot\frac{d\Vect{m_{j+1}}}{d\Vect{u_{k}}} \\
&\frac{d\Vect{st_{j+1}}}{d\Vect{u_k}} =\frac{\partial\Vect{st_{j+1}}}{\partial\Vect{a_{j}}}\cdot\frac{d\Vect{m_{j+1}}}{d\Vect{u_{k}}} \cdot \frac{1}{\delta t}\\
&\frac{d\Vect{m_{j+1}}}{d\Vect{u_{k}}} = \Big(\Vect{I}+
\frac{\partial\Vect{a_j}}{\partial\Vect{m_{j}}}\Big)\cdot 
\frac{d\Vect{m_j}}{d\Vect{u_{k}}}
\end{split}
\label{eqn:recurrenceRelation}
\end{align}

Finally gradients $\frac{d\mathcal{O}}{d\Vect{u_i}}, i=1,2,3 \cdots n_t-1$ and $\frac{d\mathcal{O}}{d\Vect{dp}}$ can be calculated by substituting the sensitivities $\frac{d\Vect{st_j}}{d\Vect{u_i}}$, $\frac{d\Vect{m_j}}{d\Vect{u_i}}$ and $\frac{d\Vect{st_j}}{d\Vect{dp}}$ in the Eqn.~\ref{eqn:finalGradient}.
\subsubsection{Residual constraint energies}


The corresponding gradient of the residual constraint energy $E_{rj} = E_j(\Vect{m_{j+1}}, \Vect{st_{j+1}}, \Vect{dp}$) is needed in order to calculate the gradient in the Eqn.~\ref{eqn:finalGradient}.
\begin{align*}
&\frac{\partial E_{rj}}{\partial\Vect{u_i}}=0; 
\quad \quad
\frac{\partial Er_j}{\partial\Vect{st_{j+1}}} = 0 \quad \text{upon convergence in Eqn.~\ref{eqn:nextState2};}\\ 
&\frac{dEr_j}{d\Vect{u_i}} = 
\frac{\partial Er_j}{\partial\Vect{u_i}}+
\frac{\partial Er_j}{\partial\Vect{m_{j+1}}}
\cdot \frac{d \Vect{m_{j+1}}}{d \Vect{u_i}}
+ \frac{\partial Er_j}{\partial\Vect{st_{j+1}}}
\cdot \frac{d \Vect{st_{j+1}}}{d \Vect{u_i}} \\
& \frac{dEr_j}{d\Vect{u_i}} = \frac{\partial Er_j}{\partial\Vect{m_{j+1}}}
\cdot \frac{d \Vect{m_{j+1}}}{d \Vect{u_i}}\\
&\frac{dEr_j}{d\Vect{dp}} = 
\frac{\partial Er_j}{\partial\Vect{dp}}+
\frac{\partial Er_j}{\partial\Vect{m_{j+1}}}
\cdot \frac{d \Vect{m_{j+1}}}{d \Vect{dp}}
+ \frac{\partial Er_j}{\partial\Vect{st_{j+1}}}
\cdot \frac{d \Vect{st_{j+1}}}{d \Vect{dp}} \\
& \frac{dEr_j}{d\Vect{dp}} = \frac{\partial Er_j}{\partial\Vect{dp}}
\end{align*}
$\frac{\partial Er_j}{\partial\Vect{m_{j+1}}}$ and $\frac{\partial Er_j}{\partial\Vect{dp}}$ is evaluated analytically and the $\frac{d \Vect{m_{j+1}}}{d \Vect{u_i}}$ is known analytically from the recurrence relation developed in Eqn.~\ref{eqn:recurrenceRelation}. 



%
\subsubsection{Updating $\Vect{u}$ and $\Vect{dp}$ }
Let $\Vect{udp}$ be the optimization variable obtained after stacking $\Vect{u}$ and $\Vect{dp}$. The gradients $\frac{d\mathcal{O}}{d\Vect{u}}$ and $\frac{d\mathcal{O}}{d\Vect{dp}}$ can be stacked to obtain the gradient $\frac{d\mathcal{O}}{d\Vect{udp}}$.
This gradient can then be used to update the optimization variable $\mathbf{udp}$ in the current step of the trajectory optimization as follows:
\begin{equation*}
\mathbf{udp_{z+1}}=\Vect{udp_{z}} - \Vect{\hat{H}}^{-1}\cdot
\evalat[\Bigg]{\frac{d\mathcal{O}}{d\mathbf{udp}}}{\Vect{udp} = \Vect{udp_z}}
\end{equation*}
We approximate $\hat{H}^{-1}$ using the L-BFGS quasi-Newton method. For standalone trajectory optimization only control inputs $\Vect{u}$ are optimized over a trajectory and design parameters $\Vect{dp}$ are kept fixed. For concurrent trajectory optimization both $\Vect{u}$ and $\Vect{dp}$ are optimized for prescribed objectives.

\section{Simulation results}
\label{sec:simResults}

\subsection{Standalone trajectory optimization}
\label{sec:controlOpt}
We start with standalone trajectory optimization in which only control inputs $\Vect{u}$ are optimized over a trajectory and design parameters $\Vect{dp}$ are kept fixed. In this subsection, we will discuss the core tasks $\mathcal{O}_{task_i}$ in the objective function $\mathcal{O}$, in Eqn.~\ref{eqn:objectiveFunction2}. Please also refer to the accompanying video.

\subsubsection{Final end-effector state goal}
To demonstrate this task, we will use the $6$-DOF simulation prototype shown in Fig.~\ref{fig:concepCCMA}(b). This prototype has a combination of actuated joints, passive joints and mobile bases.
The task, in the equation below, refers to the end-effector positioning of the CCMA system. $\Vect{X^{*}_{n_t-1}}$ specifies the final end-effector pose, where as $\Vect{X_{n_t-1}}$ is the state of the CCMA system end-effector in the final state of the trajectory. $\Vect{X_{n_t-1}}$ is a function of $\Vect{st}$.
\begin{equation}
\mathcal{O}_{task} = \mathcal{O_{\text{ee}}(\Vect{st})}
=\frac{1}{2}\norm{(\Vect{X_{n_t-1}}-\Vect{X^{*}_{n_t-1}})}_2^2
\end{equation}
\begin{figure}[!b] 
  \centering
    \includegraphics[width=0.9\columnwidth]{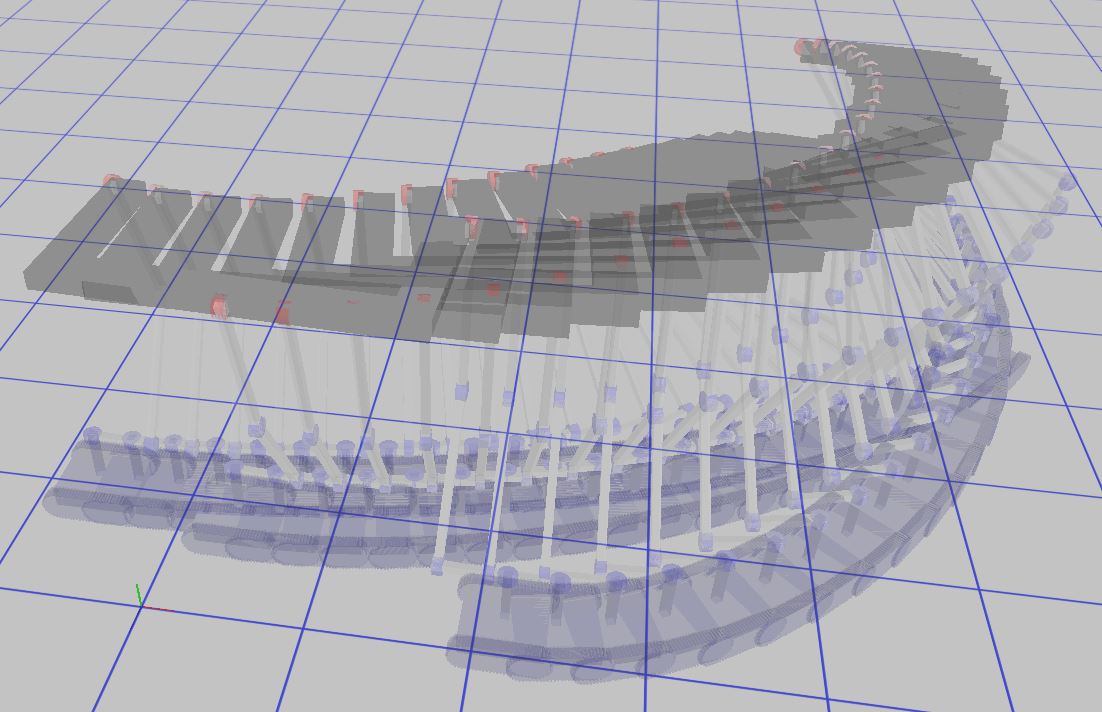}
    \caption{A snapshot of $20$ states in the trajectory for a translation plus rotation task of 6-DOF CCMA end-effector.}
    \label{fig:fig1bEETask}
\end{figure}
Fig.~\ref{fig:fig1bEETask} shows the final trajectory obtained for the task of translating the end-effector by $-3$~m along $\Vect{x_g}$, $3$~m along $\Vect{y_g}$ and $-0.5$~m along $\Vect{z_g}$ with a rotation of about $+90$~deg about the $\Vect{z_g}$. 
$n_t=20$ states were used for this simulation. The input control vector was initialized with all zero values. Therefore, initial $20$ CCMA states had the same values.

\subsubsection{Mobile agents internal collision avoidance}
While executing its tasks such as end-effector positioning, the CCMA system can have internal collision among the multiple mobile agents. In order to avoid this internal collision, we formulate a smooth continuously differentiable soft unilateral constraint $f_\text{pen}(x)$ which is defined below. Let $lim$ be the minimum distance between the mobiles bases where an internal collision is just avoided. Let $\text{dist}_m$ be the actual distance between the mobile bases. Then $x=\text{dist}_m - lim$. For $x\leq 0$, a high value of $k=10^4$ increases $f_\text{pen}(x)$ rapidly. For $0<x\leq \epsilon$, where $\epsilon$ is a safety margin over $lim$, $f_\text{pen}(x)$ increases slowly. For $x>\epsilon$, $f_\text{pen}(x)=0$.
\begin{align*}
f_{\text{pen}} (x) = \quad &k\cdot(\frac{1}{2}\cdot x^2 -\frac{\epsilon}{2}\cdot x+\frac{\epsilon^2}{6})  &x\leq 0\\
  = \quad &k\cdot(\frac{1}{6\epsilon}\cdot x^3 + \frac{1}{2} \cdot x^2 - \frac{\epsilon}{2} \cdot x + \frac{\epsilon^2}{6})   &0<x\leq \epsilon\\
  = \quad &0  &x>\epsilon
\end{align*}
$\text{dist}_m(i,j,k)$ is the distance between mobile base pair $(j,k)$ in the $i^{th}$ state of the trajectory.  This task and the corresponding objective function $\mathcal{O}_{\text{ica}}$ is added to the Eqn.~\ref{eqn:objectiveFunction2}.
\begin{multline*}
\mathcal{O}_{\text{task}} = \mathcal{O}_{\text{ica}}(\Vect{m}) = \sum_{i=0}^{n_t-1}\sum_{j=0}^{n_m-1}\sum_{k=j+1}^{n_m-1}
f_{\text{pen}}(\text{dist}_m(i,j,k))
\end{multline*}
The function $\text{dist}_m(i,j,k)$ is a function of $\Vect{m}$ and can be written as $\text{dist}_m(\Vect{m})$.
\begin{figure}[t] 
  \centering
  \vspace{7pt}
    \includegraphics[width=0.95\columnwidth]{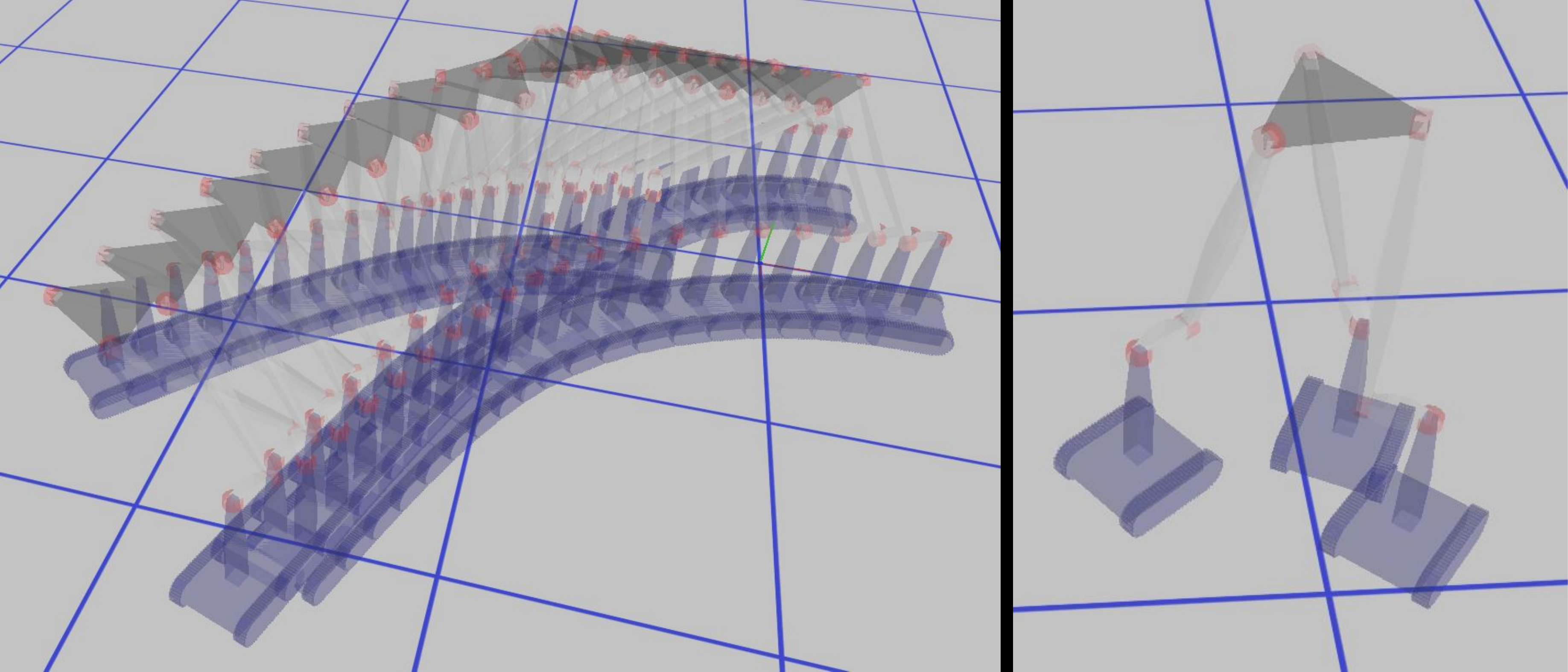}
    \caption{Without the objective $\mathcal{O}_\text{ica}$ (a) state trajectory for the end-effector positioning task (b) in the $10^{th}$ state of the trajectory, there is an internal collision between mobile bases.}
    \label{fig:ICAfigBefore}
\end{figure}

\begin{figure}[t] 
  \centering
    \includegraphics[width=0.95\columnwidth]{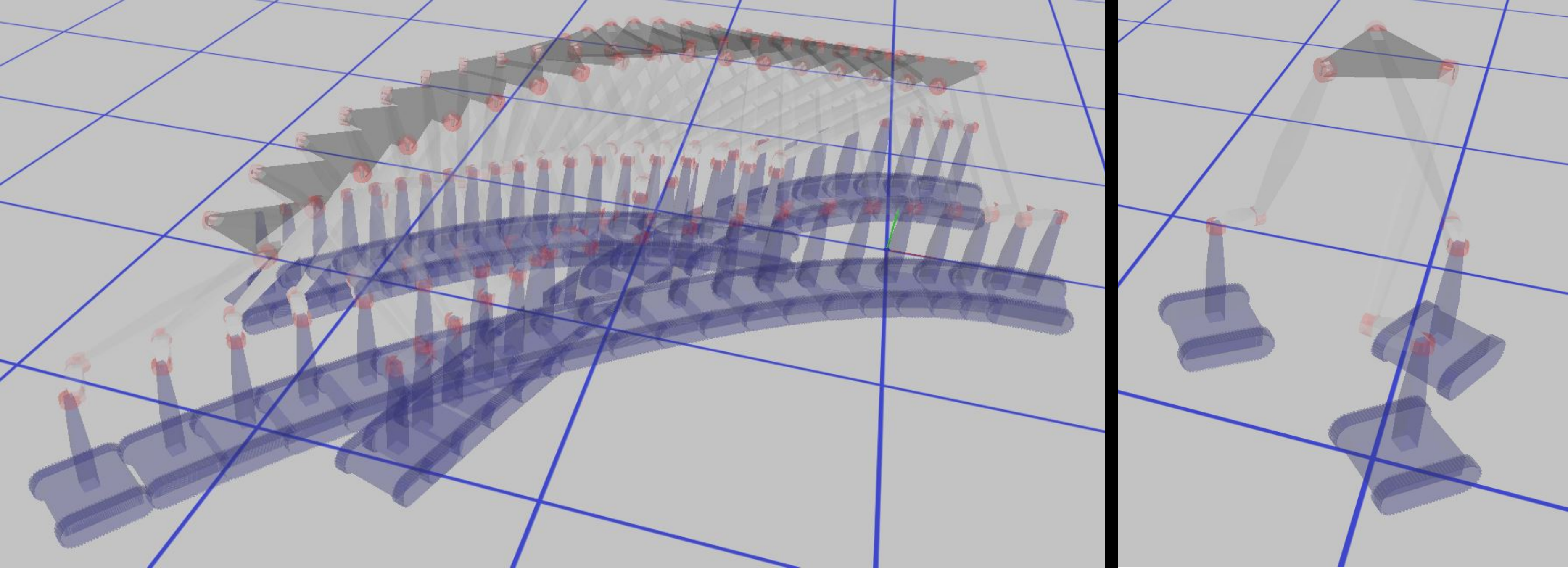}
    \caption{With the objective $\mathcal{O}_\text{ica}$ (a) state trajectory for the end-effector positioning task (b) in the $10^{th}$ state or any other state of the trajectory, there is no internal collision.}
    \label{fig:icaFigAfter}
\end{figure}

To demonstrate this task, we will use the $4$-DOF simulation prototype shown in Fig.~\ref{fig:CCMAexamples}(a). This prototype has no actuated joints. The Fig.~\ref{fig:ICAfigBefore} shows the trajectory plot obtained for the task of translating the end-effector by $-2$~m along $\Vect{x_g}$, $2$~m along $\Vect{y_g}$ with a rotation of about $90$~deg about the $\Vect{z_g}$.
In the $10^{th}$ state, there is a collision between the mobile bases. 
Once the internal collision avoidance objective is added, the internal collision is avoided among the mobile bases, as can be seen in the Fig.~\ref{fig:icaFigAfter}. 

\subsubsection{External obstacle avoidance}
\begin{figure}[t] 
  \centering
  \vspace{7pt}
    \includegraphics[width=0.95\columnwidth]{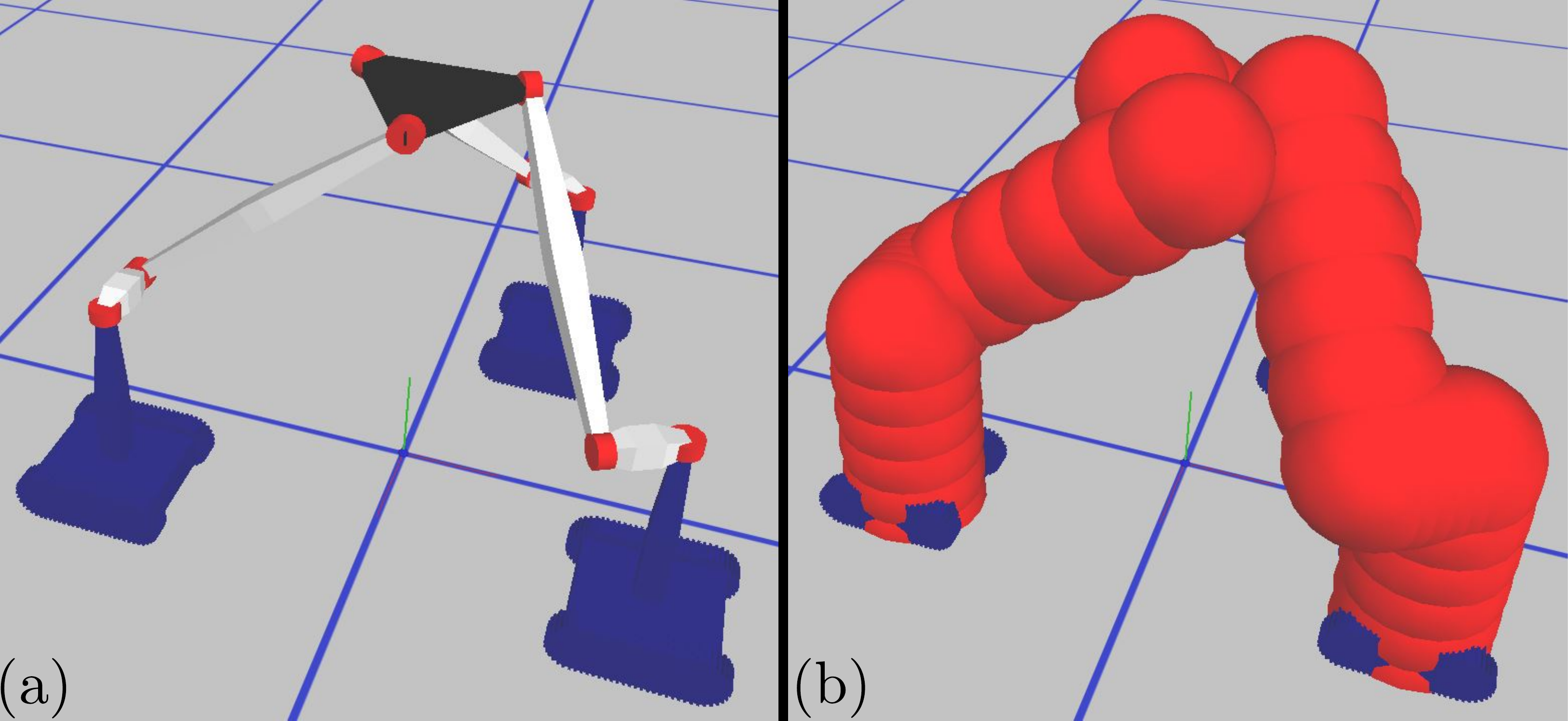}
    \caption{Side by side views of (a) $4$-DOF CCMA (b) with rigid bodies discretized into spheres.}
    \label{fig:spheresDiscMech}
\end{figure}

\begin{figure}[t] 
  \centering
    \includegraphics[width=0.95\columnwidth]{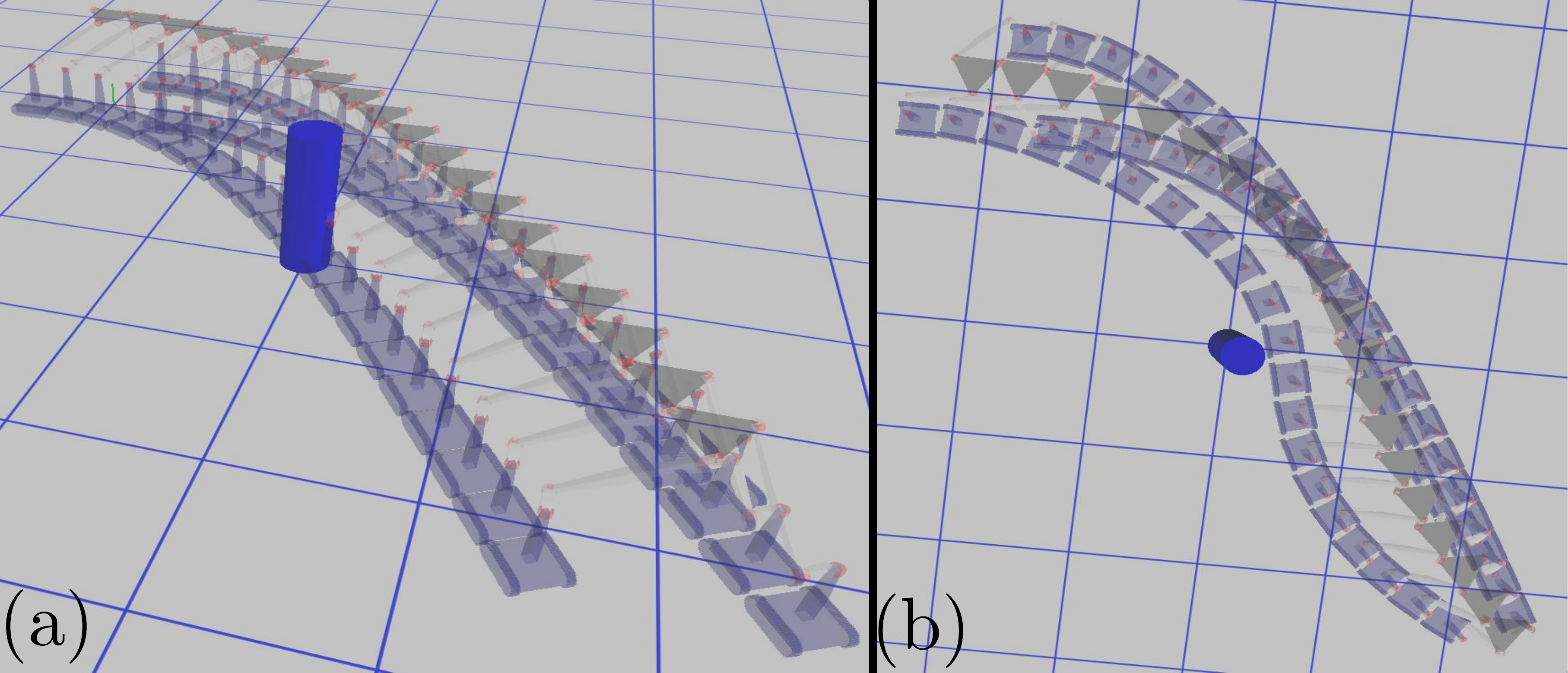}
    \caption{(a) Without $\mathcal{O}_{\text{eoa}}$,  collision with a static object (b) with $\mathcal{O}_{\text{eoa}}$, collision with a static object is avoided.}
    \label{fig:CCMAEOA}
\end{figure}

We discretize each rigid body in the CCMA system into spheres of radius $r$ with $x,y,z$ co-ordinates of its center (Fig.~\ref{fig:spheresDiscMech}). 
In order to avoid static objects (considering cylindrical primitives, x, y co-ordinates and a radius), we formulate a soft unilateral constraint $f_{\text{pen}}(\text{dist}_O)$. 
$\text{dist}_O(i,j,k)$ is the distance between sphere $j$ and the static object $k$ in the $i^{th}$ state of the trajectory.  This task and the corresponding objective function $\mathcal{O}_{\text{eoa}}$ is added to the Eqn.~\ref{eqn:objectiveFunction2}. $n_{eo}$ is the number of static objects and  $n_{s}$ is the number of spheres in the discretized CCMA system.
\begin{multline*}
\mathcal{O}_{\text{task}} = \mathcal{O}_{\text{eoa}}(\Vect{m}, \Vect{st}) = \sum_{i=0}^{n_t-1}\sum_{j=0}^{n_s-1} \sum_{k=0}^{n_{eo}-1}
f_{pen}(\text{dist}_O(i,j,k))
\end{multline*}
The function $\text{dist}_O(i,j,k)$ is a function of $\Vect{m}$, $\Vect{st}$ and can be written as $\text{dist}_O(\Vect{m}, \Vect{st})$. As shown in Fig.~\ref{fig:CCMAEOA}, objective $\mathcal{O}_{\text{eoa}}$ helps avoid collision with the static object.

\subsubsection{Smooth changes in control inputs $\Vect{u}$}
We also add a objective which tries to minimizes the change in control inputs to have smoother trajectories. 
\begin{multline*}
\mathcal{O}_{\text{task}} = \mathcal{O}_{\text{acc}}(\Vect{u}) = \frac{1}{2} \norm{\frac{\Vect{u_0}}{\delta t}}_2^2 + \sum_{i=0}^{n_t-3}\frac{1}{2} \norm{\frac{(\Vect{u_{i+1}} - \Vect{u_i})}{\delta t}}_2^2
\end{multline*} 

\subsection{Concurrent control $\Vect{u}$ and design $\Vect{dp}$ optimization} 
\label{sec:designCOncOpt}
\begin{figure}[t] 
  \centering
  \vspace{7pt}
    \includegraphics[width=1.0\columnwidth]{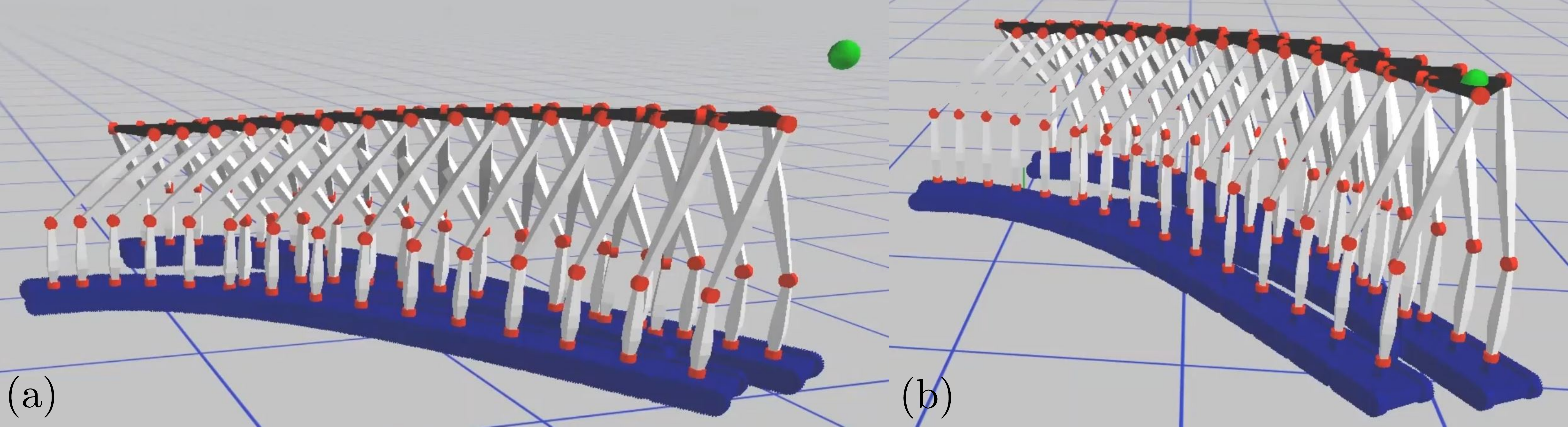}
    \caption{(a) without and (b) with concurrent design and control policy evolution during trajectory optimization.}
    \label{fig:concurrentDesignOpt}
\end{figure}

$\mathcal{O}_{ee}$, $\mathcal{O}_{eoa}$ and $Er_j$ are all functions of design parameters $\Vect{dp}$ as well. $\mathcal{O}_{ee}$ task is a classical example of how the workspace of a system is affected by design parameters. In order to demonstrate concurrent optimization of control $\Vect{u}$ and design $\Vect{dp}$ parameters, we first obtain an initial trajectory for $\mathcal{O}_{ee}$ task which is impossible to achieve with initial fixed design parameters. In second step, we do concurrent optimization but initialize the trajectory with previously calculated $\Vect{u}$ and initial design parameters. The final result  is calculation of both optimal design and control parameters which concurrently evolve for satisfying the new workspace requirements, as demonstrated in Fig.~\ref{fig:concurrentDesignOpt}. Please also refer to the accompanying video. 

\section{Experimental results}
\label{sec:expResults}
\begin{figure}[b] 
  \centering
    \includegraphics[width=0.95\columnwidth]{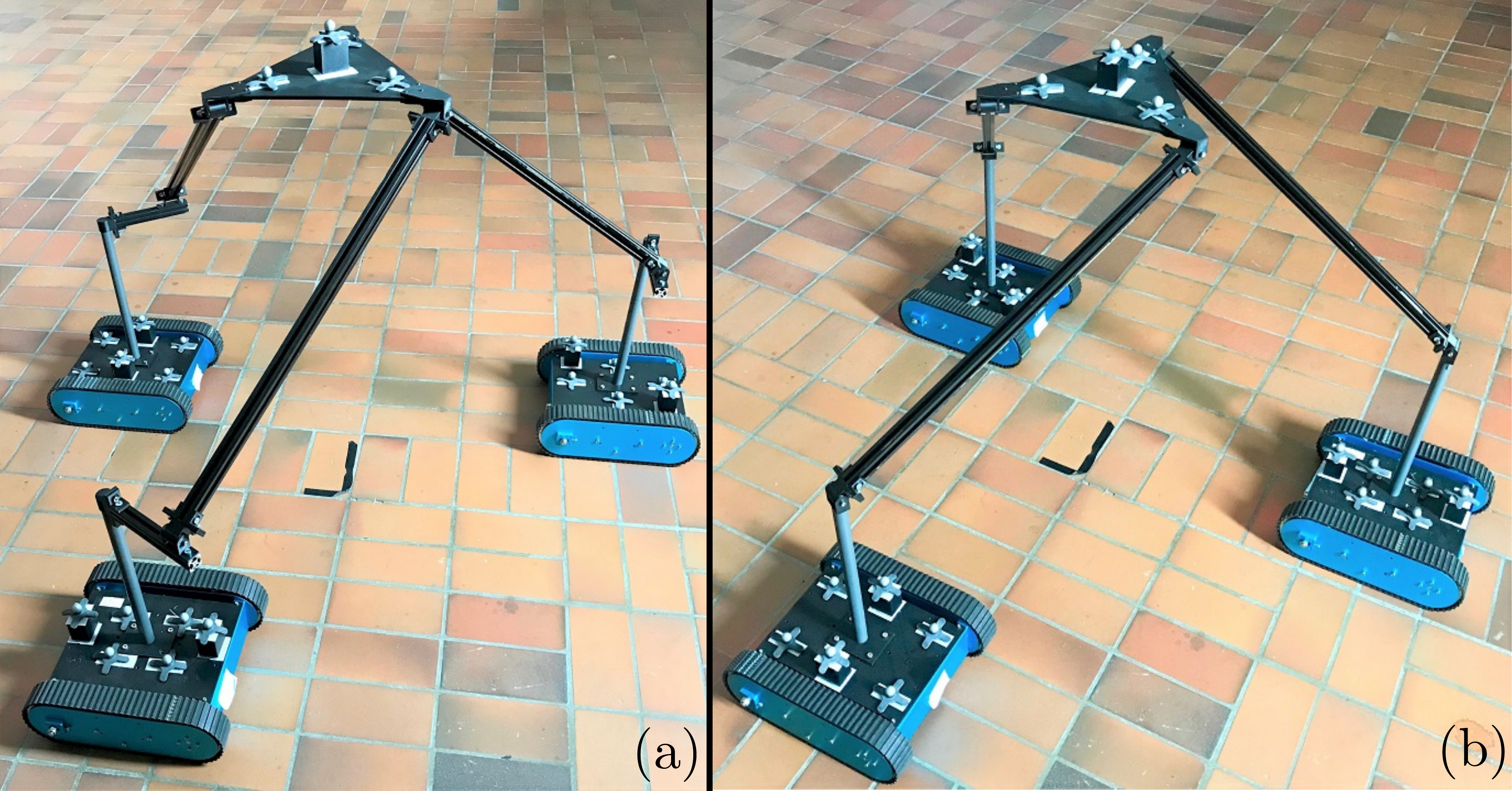}
    \caption{The fabricated physical prototype in (a) corresponds to the simulated prototype in Fig.~\ref{fig:CCMAexamples}(a) which was also used for discussing different tasks $\mathcal{O}_{task}$ of the objective function $\mathcal{O}$ in Sec.~\ref{sec:simResults}. The physical prototype in Fig.~\ref{fig:CCMAProtos}(b) has no offsets and it is a variant of the 4-DOF physical prototype in Fig.~\ref{fig:CCMAProtos}(a).}
    \label{fig:CCMAProtos}
\end{figure}
\begin{figure}[!b] 
  \centering
    \includegraphics[width=0.95\columnwidth]{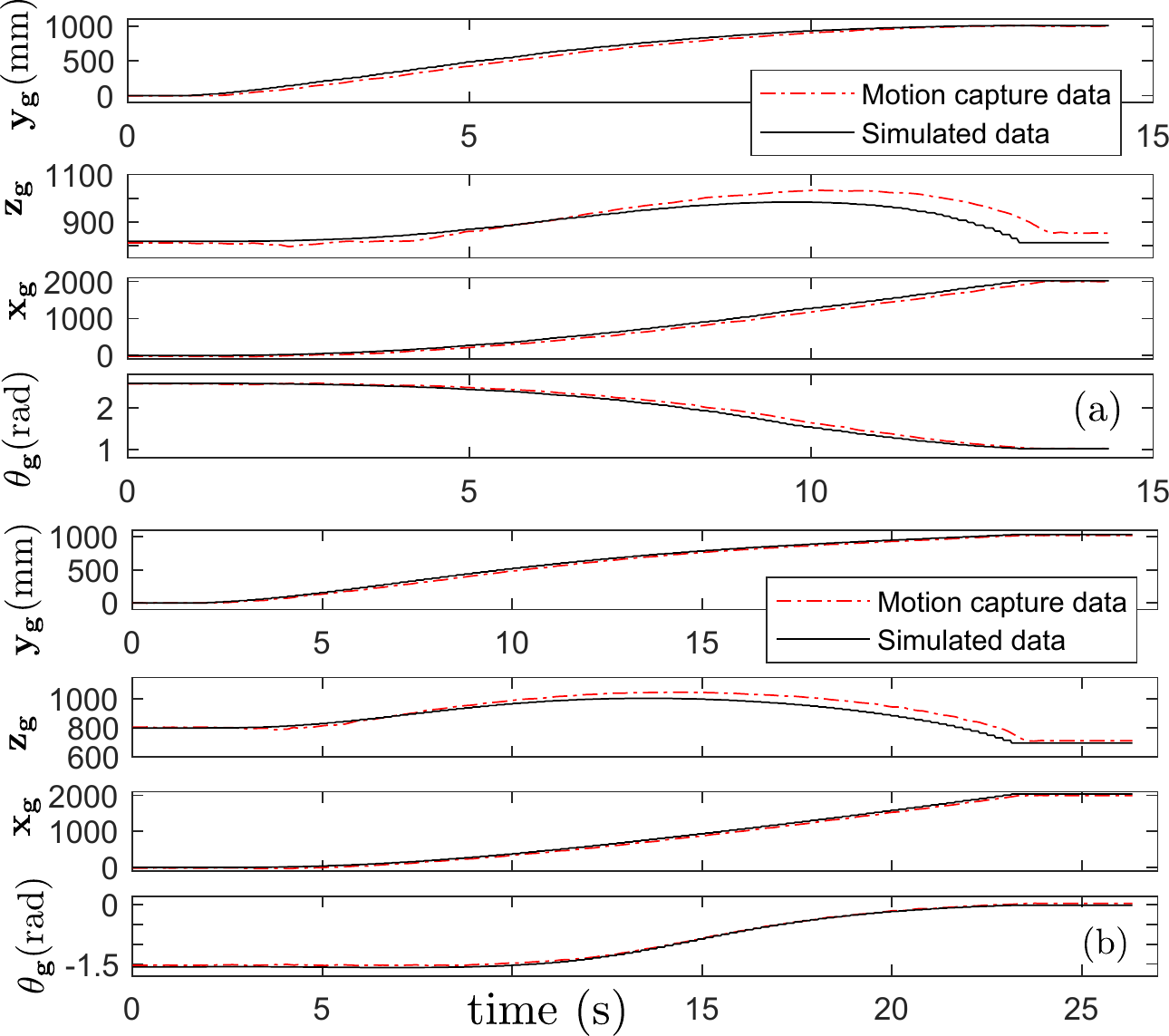}
    \caption{Tracked and simulated motion of end-effector for prototype (a) in Fig.~\ref{fig:CCMAProtos}(a))
    (b) in Fig.~\ref{fig:CCMAProtos}(b) with a simultaneous translation along $\Vect{x_g}$, $\Vect{y_g}$, $\Vect{z_g}$ and rotation about $\Vect{z_g}$.}
    \label{fig:figEE}
\end{figure}
In this section, we describe the experimental results which demonstrate the transfer of the simulation results for standalone trajectory optimization onto physical prototypes. Two fabricated physical prototypes are shown in the Fig.~\ref{fig:CCMAProtos}. For the end-effector motion shown in Fig.~\ref{fig:figEE}(a) \& (b), the physical prototype in Fig.~\ref{fig:CCMAProtos}(a) \& (b) were controlled to move $2$~m along $\Vect{x_g}$, $1$~m along $\Vect{z_g}$ with a rotation of $-90$~deg about $\Vect{z_g}$ and $2$~m along $\Vect{x_g}$, $0.1$~m along $\Vect{y_g}$, $1$~m along $\Vect{z_g}$ with a rotation of $+90$~deg about $\Vect{z_g}$, respectively. The RMSE (Root Mean Square Error) were 0.038~m (along $\Vect{y_g}$), 0.040~m (along $\Vect{z_g}$), 0.072~m (along $\Vect{x_g}$) and 0.062 rad (about $\Vect{z_g}$) for plots in Fig.~\ref{fig:CCMAProtos}(a). RMSE (Root Mean Square Error) were 0.025~m (along $\Vect{y_g}$), 0.038~m (along $\Vect{z_g}$), 0.048~m (along $\Vect{x_g}$) and 0.04 rad (about $\Vect{z_g}$) for plots in Fig.~\ref{fig:CCMAProtos}(b). Please also refer to the accompanying video. 

\section{Discussion, conclusions and future work}
\label{sec:final-sec}

\subsection{Discussion}
\label{sec:discussion}
In the Sec.~\ref{sec:controlOpt}, the initialization of control inputs is trivial ($\Vect{u}=\Vect{0}$) and a better initialization from a previous subset of solutions, like done in Sec.~\ref{sec:designCOncOpt}, or from another method like RRT (rapidly exploring random tree) could be used. The initial solution used to warm start the trajectory optimization influences convergence and the final solution. 

In each simulation example (Sec.~\ref{sec:simResults}), the orientation of the mobile bases in $\Vect{st_0}$ is trivial ($\theta^m_{0,k} = 0, \quad k=0,1,2\cdots n_m-1$). Assuming that the non-holonomic mobile bases can rotate and assume any starting orientation, one could add $\theta^m_{0,k}$ to $\Vect{udp}$ for trajectory optimization. This would significantly increase the space of feasible solutions.

The trajectory optimization method discussed in the paper focussed on CCMA systems having mobile bases with non-holonomic constraints. It is, however, readily applicable to CCMA systems with a heterogeneous combination of omni-directional bases and mobile bases with non-holonomic constraints.  

Actuating a subset of passive joints, for example in the Fig.~\ref{fig:concepCCMA}(a), has no effect on the motion generation capability if the mobile bases can fully constrain the CCMA system. But it can significantly enhance the stiffness and payload capacity of the system. However, for certain CCMA systems, for example in Fig.~\ref{fig:concepCCMA}(b), a set of actuated joints along with mobile bases are required to fully constrain them. In such cases, actuated joints are required for both  motion generation and enhancing force transmission capability.

\subsection{Conclusions and future work}
\label{sec:conclusion}
We have developed and described a novel class of mobile robotic systems having mobile bases with non-holonomic constraints, further constrained by active-passive closed-loop kinematic chains manipulating an end-effector. We have also developed a novel trajectory optimization method which can handle concurrent optimization of both design and control parameters, apart from standalone control parameters optimization. Through formulation of different tasks from gross end-effector positioning, internal mobile agents collision avoidance to external obstacle collision avoidance, we have demonstrated both in simulation and experiments, how the trajectory optimization method could be useful for the class of the CCMA systems presented in this paper.

In our future work, we would like to explore motion generation with heterogeneous combination of ground mobile robots ranging from omni-directional bases, mobile bases with non-holonomic constraints, quadruped robots and aerial mobile bases such as quadcoptors. Going further, we will also explore the force transmission capabilities, concurrently with motion generation capabilities. 










\bibliographystyle{myIEEEtran}

\bibliography{biblio-RSS}

\end{document}